\documentclass[11pt,a4paper]{article}
\usepackage[]{acl}

\usepackage{times}
\usepackage{latexsym}
\usepackage{url}
\usepackage{graphicx}
\usepackage{tabularx}
\usepackage{soul}
\usepackage{xcolor}
\usepackage{booktabs}
\usepackage{paralist} %compactitem
\usepackage{enumerate}
\usepackage[normalem]{ulem}  % for underlining words
\usepackage{epstopdf}
\usepackage{xstring}

\usepackage{color}
\usepackage{xcolor}
\newcommand{\bl}[1]{{\color{black} #1}}

\title{Breaking Bad: Norms for Valence, Arousal, and Dominance\\ for over 10k English Multiword Expressions}

 \author{Saif M. Mohammad  \\
   National Research Council Canada \\
   {\tt saif.mohammad@nrc-cnrc.gc.ca} }

\date{}

\begin{document}
\maketitle
\begin{abstract}
Factor analysis studies have shown that the primary dimensions of word meaning 
are \textit{Valence (V)}, \textit{Arousal (A)}, and \textit{Dominance (D)}. 
Existing lexicons such as the NRC VAD Lexicon, published in 2018, include VAD association ratings for \textit{words}.
Here, we present a complement to it, which has human ratings of valence, arousal, and dominance for $\sim$10k
English Multiword Expressions (MWEs) and their constituent words. 
% We also include entries for 
\bl{We also increase the coverage of unigrams, especially}
words that have become more common since 2018.
% that we are not included in NRC VAD. 
In all, %our VAD Lexicon has 
\bl{the new NRC VAD Lexicon v2 now has}
% we have added to the NRC VAD Lexicon}
entries for $\sim$10k MWEs and $\sim$25k words,
\bl{in addition to the entries in v1.}
% \bl{We refer to this lexicon as NRC VAD Lexicon v2.}
We show that the associations are highly reliable.
We use the lexicon to examine emotional \bl{characteristics} of MWEs, including: 1. The 
\bl{degree to which MWEs (idioms, noun compounds, and verb particle constructions) exhibit strong emotionality;} %Similar analysis is done for arousal and dominance. 
2. The degree of emotional compositionality in MWEs.
The lexicon enables a wide variety of   
research in NLP, Psychology, Public Health, Digital Humanities, and Social Sciences. 
\bl{The NRC VAD Lexicon v2 is freely available
through the project webpage:}  \url{http://saifmohammad.com/WebPages/nrc-vad.html}
\end{abstract}

\section{Introduction}

Several influential factor analysis studies have shown 
that the three most important, largely independent, dimensions of connotative meaning and emotions are
valence (positiveness--negativeness/pleasure--displeasure), arousal (active--passive), and dominance %aka competenece 
(dominant--submissive, competent--incompetent, powerful--weak) % in control--out of control)
\cite{Osgood1957,russell1980circumplex,russell2003core}. We will refer to the three dimensions individually as {\it V, A,} and \textit{D}, and together as \textit{VAD}.
Language is a powerful medium for expressing emotions (consciously and unconsciously) and language-resource work has produced large repositories of word--emotion associations and sentences annotated for emotions. However, we are not aware of any large scale work on multiword expressions (MWEs) and VAD. 

MWEs have been defined with some differences in past works, but here we simply consider sequences of two or more words (often with some interesting semantic, syntactic, or functional property) as MWEs. 
Broadly speaking, MWEs are important in NLP, linguistics, social sciences, and psychology because their meaning is often not compositional \cite{smolka2020role} and MWEs reveal insights about the structure of language, social interaction, and cognitive processing. Yet, unlike their lexical or sentence cousins, far fewer language resources exist for MWEs.

Our work at the intersection of emotions and MWEs makes these contributions:
% \begin{compactenum}
\begin{enumerate}
    \item We obtained human ratings of valence, arousal, and dominance for about 10,000 common English MWEs. 
    \item We also obtained VAD ratings for about 25,000 English words that are not included in the NRC VAD Lexicon v1 \cite{mohammad-2018-obtaining}. These include terms that have become more common since 2018 (such as \textit{quarantine)} as well as words that are constituents of the MWEs included in the lexicon.
\item The scores are fine-grained real-valued numbers from -1 (lowest V, A, or D) to 1 (highest V, A, or D). We show that the annotations lead to reliable VAD score (split-half reliability scores of  $r=0.99$ for valence, $r=0.98$ for arousal, and $r=0.96$ for dominance.) 
We will refer to this lexicon as {\it MWE-VAD}.  
\bl{The new MWE and unigram annotations are added to the entries in NRC VAD Lexicon v1 to form the NRC VAD Lexicon v2.}

    \item We use the \bl{newly created} lexicon to examine emotionality of MWEs, including: 
    \begin{enumerate}
        
    \item The distributions of high- and low-valence MWEs in different types of MWEs such as idioms, noun compounds, and verb particle constructions. Similar analysis is done for arousal and dominance. 
    \item  The degree of emotional compositionality in MWEs --- i.e., to what extent the emotionality of an MWE can be determined from the emotionality of its constituent words?
    \end{enumerate}
\item Finally, we describe a number of research and application directions that can benefit from MWE-VAD. 
 One can use MWE-VAD to study MWEs specifically
\bl{and NRC VAD v2} for work on valence, arousal, and dominance, in general. \bl{NRC VAD Lexicon v2 (along with automatic  translations of the English terms to over 100 languages) is made freely available for research 
through the project webpage.}\footnote{\textbf{VAD v2:} \url{http://saifmohammad.com/WebPages/nrc-vad.html}\\
\textbf{Emotion Dynamics Code} \cite{vishnubhotla-mohammad-2022-tusc} to analyze emotions in text using emotion lexicons: \url{https://github.com/Priya22/EmotionDynamics}.}
\end{enumerate}
%\end{compactenum}

All of the annotation tasks described in this paper were approved by 
our institution's review board, which examined the methods to ensure that they were ethical. Special attention was paid to obtaining informed consent and protecting participant anonymity.

\section{Related Work}

\noindent {\bf Primary Dimensions of Meaning and Affect.} Highly influential, psycholinguistics and affective science work by \newcite{Osgood1957} and \newcite{russell1980circumplex}, respectively, asked human participants to rate words along dimensions
of opposites such as {\it heavy--light, good--bad, strong--weak,} etc.
Factor analysis of these judgments revealed 
that the three most prominent dimensions of connotative meaning and emotion
are valence ({\it pleasure--displeasure}), arousal ({\it active--passive}), and dominance ({\it strong--weak}).\\[5pt] 
\noindent {\bf MWEs.} MWE work in NLP has focused on the automatic discovery, processing, and understanding of MWEs from corpora (see surveys such as \newcite{smolka2020role} and \newcite{constant-etal-2017-survey}). 
Less work has gone into manually compiling lists of MWEs. Notably,
\newcite{PMID:35867207} manually compiled a lexicon of about 62,000 MWEs. 
They added concreteness ratings for the MWEs, as well as the frequencies of the MWEs in a subtitles corpus \cite{brysbaert2012adding}.
\newcite{takahashi2024comprehensive} compiled a lexicon of 160,000 Japanese MWEs.
\newcite{tong2024metaphor} compiled a 10k English metaphor--literal paraphrase pairs dataset.
There is even less work on annotating MWEs for emotions, despite work on many *word*--emotion association lexicons.
\newcite{jochim-etal-2018-slide} crowdsourced the sentiment annotation of 5000 English idioms.
\newcite{ibrahim2015idioms} annotated 3600 Modern Standard Arabic idioms for sentiment.\\[5pt]
\noindent {\bf Existing Affect Lexicons.} 
The \newcite{bradley1999affective} 
lexicon has more than 1000 words 
with real-valued scores of 
valence, arousal, and dominance. 
 For each word, they
 asked annotators to rate valence, arousal, and dominance---for more than 1,000 words---on a 9-point rating scale. The ratings from multiple annotators were averaged to obtain a score between 1 (lowest V, A, or D) to 9 (highest V, A, or D).
Their lexicon, called the {\it Affective Norms of English Words (ANEW)}, has since been widely used across many different fields of study.
ANEW was also translated into non-English languages: e.g., \newcite{moors2013norms} for Dutch,  \newcite{vo2009berlin} for German, and \newcite{redondo2007spanish} for Spanish.
\newcite{warriner2013norms} created a VAD lexicon for more than 13,000 words, using a similar annotation method as for ANEW. 
The NRC VAD Lexicon v1 \cite{mohammad-2018-obtaining} is the largest manually created VAD lexicon (in any language).
It has entries for about 20,000 English words.
The NRC Emotion Lexicon was created by crowdsourcing and it includes association entries for
 about 14,000 words with eight Plutchik emotions as well as positive and negative sentiment 
\cite{MohammadT13,MohammadT10}.\footnote{\url{http://saifmohammad.com/WebPages/NRC-Emotion-Lexicon.htm}}

\bl{The 10K MWEs and many unigrams from VAD v2 have since been used in other annotation projects as well. The NRC WorryWords Lexicon \cite{worrywords-emnlp2024} has real-valued scores indicating their associations with anxiety for roughly the same 44k English words and 10k MWEs. 
The non-neutral subset of the VAD v2 (those with valence scores lower or equal to $-$0.33 and those with valence scores higher or equal to 0.33) were used to create the NRC Words of Warmth (WoW) Lexicon. WoW is a list of about 31,000 English terms ($\sim$26k unigrams and $\sim$5k MWEs) and real-valued scores indicating their associations with warmth, sociability, and trust---core dimensions (along with dominance aka competence) in social cognition and stereotypes \cite{fiske2002,bodenhausen2012social,fiske2018stereotype,abele2016facets,koch2024validating}.
}\\[5pt]
\noindent {\bf Automatically Generated Affect Lexicons.} There is a large body of work on automatically determining word--sentiment, word--emotion, and word--VAD associations, including:
\newcite{StrapparavaV04,yang2007building,Mohammad12,COIN:COIN12024,yu2015predicting,staiano2014depechemood,bandhakavi2021emotion,muhammad-etal-2023-afrisenti} to name just a few.  These methods, \bl{including those that employ large language models,}  often assign a real-valued score  representing the degree of association. 
Our MWE-VAD Lexicon can enable further such work on MWEs, especially by keeping a portion as a 
\bl{source of seed/example words 
for training/few-shot learning} 
and a held-out portion for evaluating the automatically generated lexicons.

\section{Obtaining Human Ratings of Valence, Arousal, and Dominance}
The keys steps in obtaining the new annotations were as follows:\\[-10pt] 
\begin{compactenum}
\item selecting the terms to be annotated 
\item developing the questionnaire
\item developing measures for quality control (QC)
\item annotating terms on a crowdsource platform
\item discarding data from outlier annotators (QC)
\item aggregating data from multiple annotators to determine the VAD association scores
% for each of the terms. 
\end{compactenum}
We describe each of the steps below.\\[5pt]
\noindent {\bf 1. Term Selection.}
We wanted to include various kinds of multi-word expressions, including common phrases, light verb constructions, idiomatic constructions, etc.
However, identifying MWEs from a large corpus of text is not trivial.
Further, we wanted to include terms for which other linguistically interesting annotations already exist (such as concreteness ratings). 
Thus, for our work we chose the 10,500 most frequent MWEs compiled by \newcite{PMID:35867207}.

The NRC VAD Lexicon v1 includes about 20,000 common English words. However, with the passage of time, words that were less prominent earlier can become more commonly used: e.g., \textit{quarantine, deepfake, lockdown, workstation,} and \textit{gaslighting}. 
Further, we wanted to include words that were constituents of the MWEs chosen for annotation. This would allow for comparisons of the VAD scores of MWEs and their constituents.
Finally, we wanted to include terms for which other linguistically interesting annotations already exists (such as concreteness and age of acquisition ratings).
Therefore we included words from the Prevalence dataset \cite{brysbaert2019word}. 
    This dataset has prevalence scores (how widely a word is known by English speakers), determined directly by asking people, for 62,000 lemmas. We included a term if it was marked as known to at least 70\% of the people who provided responses for the term. 
    From this set we removed terms that are common person names or city names; and also words already annotated for VAD in the NRC VAD lexicon.
This resulted in close to 25k unigrams that we annotated for VAD.\\[5pt]
\noindent{\bf 2. VAD Questionnaires}
The questionnaires used to annotate the data 
 were developed  after several rounds of 
 pilot annotations. 
 Detailed directions, including notes directing respondents to consider predominant word sense (in case the word is ambiguous) and example questions (with suitable responses) were provided. (See Appendix.)
 The primary instruction and the questions presented to annotators are shown below. %\\[-14pt]
% The annotation questions and the instructions for the annotators are shown in a supplementary file. 

% \newpage
{
\noindent\makebox[\linewidth]{\rule{0.48\textwidth}{0.4pt}}\\% [-8pt]
{ \small
% Summary Instructions
\noindent VALENCE: Consider positive feelings (or positive sentiment)  to be a broad category that includes:\\
\indent \textit{positiveness / pleasure / goodness / happiness /\\
\indent greatness / brilliance / superiority / health  etc.}\\
Consider negative feelings (or negative sentiment) to be a category that includes:\\
\indent \textit{negativeness / displeasure /badness / unhappiness /\\ \indent insignificance / terribleness / inferiority / sickness etc.}\\
% nonchalant, uninterested, 
 If you do not know the meaning of a word or are unsure, you can look it up in a dictionary (e.g., the Merriam Webster) or on the internet.

\noindent \textbf{Quality Control}

\noindent Some questions have pre-determined correct answers. If you mark these questions incorrectly, we will give you immediate feedback in a pop-up box. An occasional misanswer is okay. However, if the rate of misanswering is high (e.g., $>$20\%), then all of one's HITs may be rejected.

\noindent Select the options that most English speakers will agree with.\\[4pt]
\noindent Q1.  $<$term$>$ is often associated with:\\[-1pt]
\indent 3: very positive feelings\\[-1pt]
\indent 2: moderately positive feelings\\[-1pt]
\indent 1: slightly positive feelings\\[-1pt]
\indent 0: not associated with positive or negative feelings\\[-1pt]
\hspace*{-2mm} \indent -1: slightly negative feelings\\
\hspace*{-2mm} \indent -2: moderately negative feelings\\
\hspace*{-2mm} \indent -3: very negative feelings\\[-8pt]
}
\noindent\makebox[\linewidth]{\rule{0.48\textwidth}{0.4pt}}\\[-15pt]

}

{
\noindent\makebox[\linewidth]{\rule{0.48\textwidth}{0.4pt}}\\ %[3pt]
{ \small
\noindent AROUSAL: This task is about words and their association with activeness or arousal.
\noindent Consider activeness or arousal to be a broad category that includes:

    \textit{active, aroused, stimulated, frenzied, 
    excited, jittery, alert,}\\ \indent etc.

\noindent Consider inactiveness or calmness to be a broad category that includes:

    \textit{inactive, calm, unaroused, passive, relaxed, sluggish,} etc.

\noindent This task is not about sentiment. (For example, something can be positive and inactive (such as flower), positive and active (such as exercise and party), negative and active (such as murderer), and negative and inactive (such as negligent).\\[-16pt]

\noindent\makebox[\linewidth]{\rule{0.48\textwidth}{0.4pt}}\\[-12pt]

}

{
\noindent\makebox[\linewidth]{\rule{0.48\textwidth}{0.4pt}}\\% [-8pt]
{ \small
% Summary Instructions
\noindent DOMINANCE: This task is about words and their association with dominance, competence, control of situation, or powerfulness.
Consider dominance, competence, control of situation, or powerfulness to be a broad category that includes:

   \indent \textit{ dominant, competent, in control of the situation,}\\ \indent \textit{powerful, influential, important, autonomous,} etc.\\
\noindent Consider submissiveness, incompetence, controlled by outside factors, or weakness to be a broad category that includes:

    \indent \textit{submissive, incompetent, not in control of the situation,}\\
    \indent \textit{weak, influenced, cared-for, guided,} etc.\\
\noindent This task is not about sentiment. (For example, something can be positive and weak (such as a flower petal) and something can be negative and strong (such as tyrant).\\[-16pt]

% \noindent If you do not know the meaning of a word or are unsure, you can look it up in a dictionary (e.g., the Merriam Webster) or on the internet.

% \noindent A rule of thumb is that a term associated with more dominance, competence, control of situation, or powerfulness tends to often occur in sentences that convey dominance, competence, control of situation, or powerfulness, whereas a term associated with more submissiveness, incompetence, controlled by outside factors, or weakness tends to often occur in sentences that convey submissiveness, incompetence, controlled by outside factors, or weakness.

\noindent\makebox[\linewidth]{\rule{0.48\textwidth}{0.4pt}} \\[3pt]
}
% \noindent The full questionnaire will be made available on the project webpage. 
\noindent{\bf 3. Quality Control Measures.}
%brb author no s below
About 2\% of the data was annotated beforehand by the authors and interspersed with the rest. These questions are referred to as \textit{gold} (aka \textit{control}) questions. 
% During crowd annotation, 
% We interspersed the gold questions with the other questions.
% and the annotator is not aware which is which. However, 
Half of the gold questions were used to provide immediate feedback to the annotator (in the form of a pop-up on the screen) in case they mark them incorrectly. We refer to these as \textit{popup gold}. This helps prevent the situation where one annotates a large number of instances without realizing that they are doing so incorrectly. 
It is possible, 
that some annotators share answers to gold questions with each other (despite this being against the terms of annotation). 
% The gold questions also served as examples to guide the annotators.
Thus, the other half of the gold questions were also separately used to track how well an annotator was doing the task, but for these gold questions no popup was displayed in case of errors. 
We refer to these as 
\textit{no-popup gold}.\\[3pt]
% If a crowd worker answered a gold question incorrectly, then they were immediately notified.
% \sm{the annotation was discarded, and an additional annotation was requested from a different annotator}. 
\noindent{\bf 4. Crowdsourcing.} 
We setup the annotation tasks on the crowdsourcing platform, {\it Mechanical Turk}.
 In the task settings, we specified that we needed annotations from nine people for each word. 
% We obtained annotations from native speakers of English residing around the world. 
Annotators were free to provide responses to as many terms as they wished. 
The annotation task was approved by 
our institution's review board.

\noindent {\it Demographics:} Since location and culture can impact word--association norms, and because AMT workers are mostly from the US, we requested annotations from participants who live in USA and Canada. 
The average age of the respondents was 34 years. Among those that disclosed their gender, about 53\% were female, 47\% were male.\footnote{Respondents were shown optional text boxes to disclose their demographic information as they choose; especially important for social constructs such as gender, in order to give agency to the respondents and to avoid binary language.}\\[3pt]
\noindent{\bf 5. Filtering.} % and Re-annotation} %  as part of Quality Control} 
If an annotator's accuracy on the gold questions (popup or non-popup) fell below 80\%, then they were refused further annotation, 
 and all of their annotations were discarded (despite being paid for).\\[3pt]
\noindent{\bf 6. Aggregation.} 
Every response was mapped to an integer from -3 (highly negative/inactive/submissive) to 3 (highly positive/active/dominant) as follows: \\[-10pt]
 \begin{compactitem}
% \begin{itemize} 
    \item highly positive/active/dominant: 3
    \item moderately positive/active/dominant: 2
    \item slightly positive/active/dominant: 1
    \item neither positive/active/dominant nor  negative/inactive/submissive: 0
    \item slightly negative/inactive/submissive: -1
    \item moderately negative/inactive/submissive: -2
    \item highly negative/inactive/submissive: -3
% \end{itemize}
 \end{compactitem}
 \vspace*{2mm}
\noindent The final  score for each term is simply the average score it received from each of the annotators.
 The scores were then linearly transformed to the interval: -1 (highest negativeness/inactivity/submissiveness) 
 to 1 (highest positiveness/activity/dominance).
 See Table~\ref{tab:ann} for summary statistics.

\begin{table}[t!]
\begin{center}
%\vspace*{-4mm}
\small{
\begin{tabular}{lrrr}
\hline 
{\bf Version} 	             & \bf \#Words& \bf \#MWEs & \bf \#Total\\\hline
NRC VAD v1 (2018)     & 19,839       & 132     & 19,971\\
MWE VAD (2025)     & 25,089       &10,073 & 35,162\\
NRC VAD v2 (2025)     & 44,928       &10,205 & 55,133\\
\hline
\end{tabular}
}
\caption{\label{tab:ann} {Number of terms in the NRC VAD v1, MWE VAD, and the combined lexicon (NRC VAD v2).}
}
\vspace*{-3mm}
\end{center}
\end{table}

\begin{table}[t!]
\begin{center}
%\vspace*{-4mm}
\small{
\begin{tabular}{lrrr}
\hline 
{\bf Dimension} 	             & \bf Avg.\@ \#Annot. &\bf SHR ($\rho$)		 &\bf SHR ($r$) \\\hline
valence         & 7.83      &0.98   &0.99     \\
arousal         & 7.96      &0.97       &0.98     \\
dominance       & 8.06      &0.96     &0.96\\
\hline
\end{tabular}
}
%\vspace*{-2mm}
\caption{\label{tab:shr-words} {Average number of annotations per term and split half reliability measured through both Spearman rank ($\rho$) and Pearson's ($r$) correlations. Scores in the 0.9s indicate high reliability.}
}
% \vspace*{-3mm}
\end{center}
\end{table}

\section{Reliability of the Annotations} 

A useful measure of quality is the reproducibility of the end result---repeated independent manual
annotations from multiple respondents should result in similar  scores.
To assess this reproducibility, we calculate
average {\it split-half reliability (SHR)} over 1000 trials. SHR is a common way to determine reliability of responses to generate scores on an ordinal scale %in the fields of psychology and psycholinguistics 
\cite{weir2005quantifying}.} 
  All annotations for an
item are randomly split into two halves. Two separate sets of scores are aggregated, just as described in Section 3 (bullet 6), from the two halves. 
Then we determine how close the two sets of scores are (using a metric of correlation).  
This is repeated 1000 times and the correlations are averaged.
The last two columns in Table~\ref{tab:shr-words} show the results (split half-reliabilities). Spearman rank and Pearson correlation scores of over 0.95  for V, A, and D indicate high reliability of the real-valued scores obtained from the annotations. (For reference, if the annotations were random, then repeat annotations would have led to an SHR of 0. Perfectly consistent repeated annotations lead to an SHR of 1. Also, similar past work on word--anxiety associations had SHR scores in the 0.8s \cite{mohammad-2024-worrywords}.) 

\begin{figure}[t!]
	\centering
	    \includegraphics[width=0.47\textwidth]{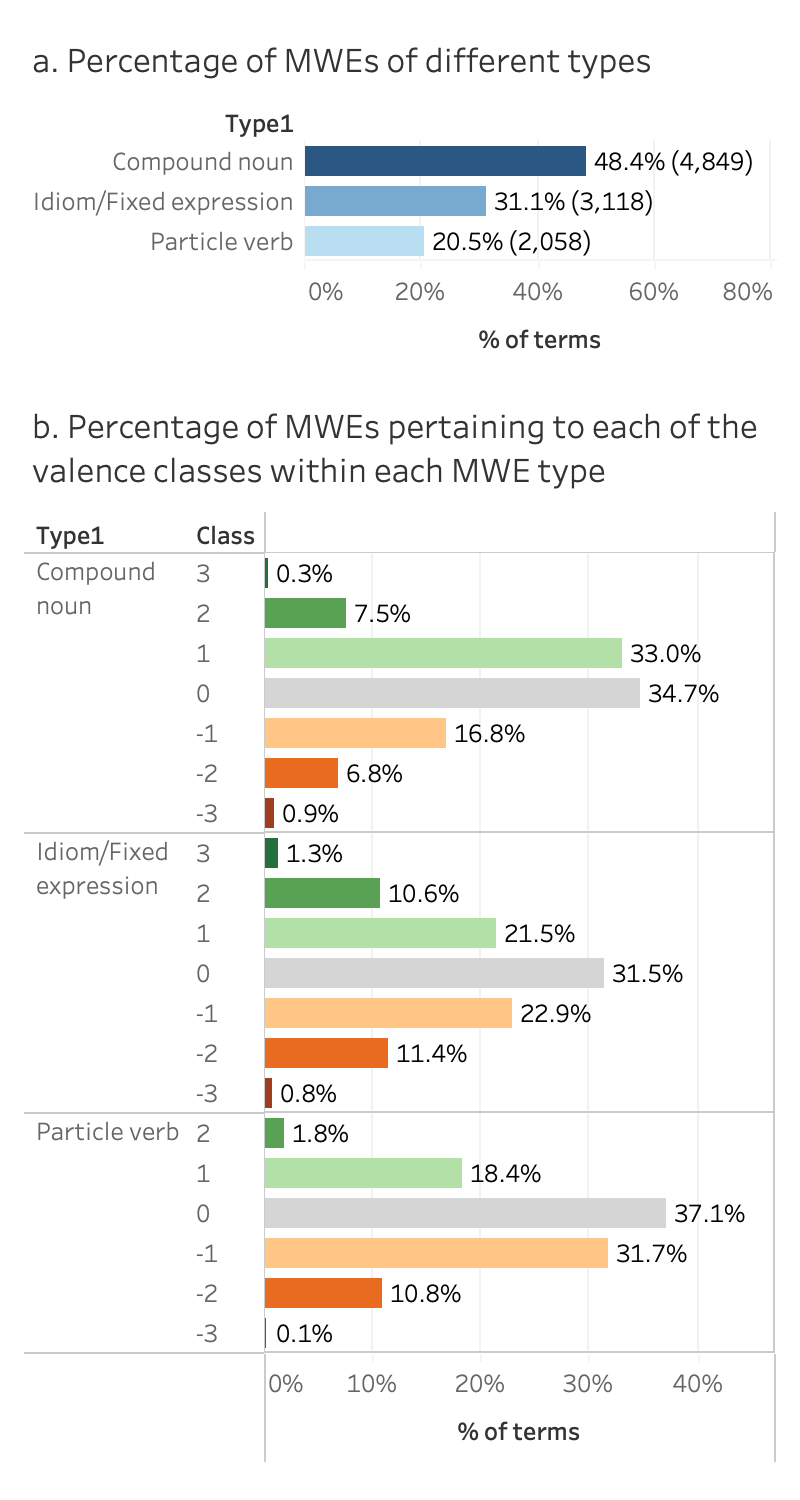}
        \caption{Distributions of MWE types.}
        % Percentage of MWEs pertaining to each of the valence classes within each MWE type.}
      %  \vspace*{-3mm}
	    \label{fig:val-distrib}
\end{figure}

\section{How Commonly do MWEs Convey Strong Emotionality?}
% Associated with High and Low VAD?}

MWEs can be of different types such as noun compounds (non--noun collocations), idioms/fixed expressions, particle verb constructions, etc. Each of these types is relevant to expressing \bl{strong emotionality (high and low VAD rather than just neutral VAD)}. For example, \textit{breath of fresh air, over the moon, kicked the bucket}, and \textit{cold shoulder} are idiomatic expressions conveying various levels of valence. Similarly, \textit{make a scene, take a leap, take a seat,} and \textit{make a wish} are light verb constructions that convey different levels of arousal.
And, \textit{power move, victory lap, support system, death spiral,} etc.\@ are noun compounds conveying various levels of dominance. Yet, we do not know the extent to which these different types of MWEs are associated with high and low VAD: e.g., how common is high dominance association in light verb constructions? Knowing these distributions will shed more light on how we use different types of MWEs to express emotions.

Each MWE entry in the MWE concreteness norms dataset \cite{muraki2023concreteness} is marked with information about its MWE type.
We make use of this to determine
 (a) percentage of different types of MWEs
--- shown in Figure \ref{fig:val-distrib} (a); and (b) the  percentage of MWEs in various valence classes within various types of MWEs --- shown in in Figure \ref{fig:val-distrib} (b). (The numbers within each type sum up to 100\%.)
Similar plots for arousal and dominance are shown in Figures \ref{fig:aro-distrib} and \ref{fig:dom-distrib}, respectively
(in the Appendix).

\noindent \textbf{Results:} From Figure \ref{fig:val-distrib} (a) we see that compound nouns are the most frequent class of MWEs in MWE-VAD ($\sim$48\%), followed by fixed expressions and then particle verbs.
Observe in \ref{fig:val-distrib} (b) that 
the percentage of non-neutral MWEs varies from about 63\% in particle verb constructions to 69\% in idioms and fixed expressions. Thus idioms seem to be particularly useful in terms of conveying valence. Further, in all three MWE types, the proportions for the low-valence (negative) classes are higher than the proportions for the high-valence (positive) classes. This is inline with what was found for words \cite{schrauf2004preponderance,mohammad-2018-obtaining}, and consistent with the hypotheses: 1. it is evolutionarily more useful to clearly identify different negative valence situations (requiring a greater vocabulary of negative words and expressions) than positive situations; and 
2. human beings spend more time and more effort in thinking about negative experiences, thereby coming up with more negative words for the more careful and detailed deliberation  \cite{schrauf2004preponderance}.

The arousal and dominance distributions (shown in the Appendix) reveal that idioms and particle verb constructions have more low arousal and low dominance MWEs (than high A/D), whereas noun compounds have markedly more high arousal and high dominance  expressions (than \bl{low} A/D). This indicates that high-dominance and high-arousal concepts are more likely to be lexicalized (i.e., turned into fixed phrases or compounds). Power-related entities (e.g., \textit{leader, boss, commander}) are culturally and cognitively salient and thus appear in compound forms more frequently: for example, \textit{power play, executive order, master plan,} etc.
High-arousal experiences more often result in visible actions, events, or consequences.
This makes them easier to name and more likely to become lexicalized as noun compounds:
% Compounds often form around discrete, nameable things (
e.g., \textit{car crash, power surge, stress test}, etc.

Overall, these results show that all three types of MWEs are substantial sources of expressing high and low VAD, and that some types of MWEs are more amenable to express some types of emotionality (e.g., noun compounds more likely to express high arousal and dominance).

\section{Emotional Compositionality of MWEs}

MWEs are especially interesting because often their meaning is noncompositional. Thus, the emotionality of the MWE may not be predictable from the emotionality of its constituent words. Yet, little is known about the extent to which this is true. Further, we do not know the extent to which neutral words come together to create high- or low-valence/arousal/dominance MWEs.

To explore this, we focused on the 8,330 bigram (two-word sequence) MWEs. We will refer to the first word in the MWE as word1 and second as word2. For each of the dimensions (V/A/D), we partitioned them into 49 (7*7) bins corresponding to every class combination of word1 and word2: high V/A/D--high V/A/D, high V/A/D--moderate V/A/D,..., neutral--neutral,..., low V/A/D--low V/A/D.
We then determined the average V/A/D scores of all the MWEs in each bin. We show the results for valence in Figure \ref{fig:v-compo} (a). For arousal and dominance the results are shown in Figures \ref{fig:a-compo} (a) and \ref{fig:d-compo} (a), respectively, in the Appendix.
For each of the 49 bins, %apart from the average anxiety score, 
we also computed the percentages of bigram MWEs associated with high V/A/D (Figures \ref{fig:v-compo} (b), \ref{fig:a-compo} (b), \ref{fig:d-compo} (b)) and the percentages of bigram MWEs associated with low V/A/D (Figures \ref{fig:v-compo} (c), \ref{fig:a-compo} (c), \ref{fig:d-compo} c)).

\begin{figure*}[t!]
	\centering
	    \includegraphics[width=0.84\textwidth]{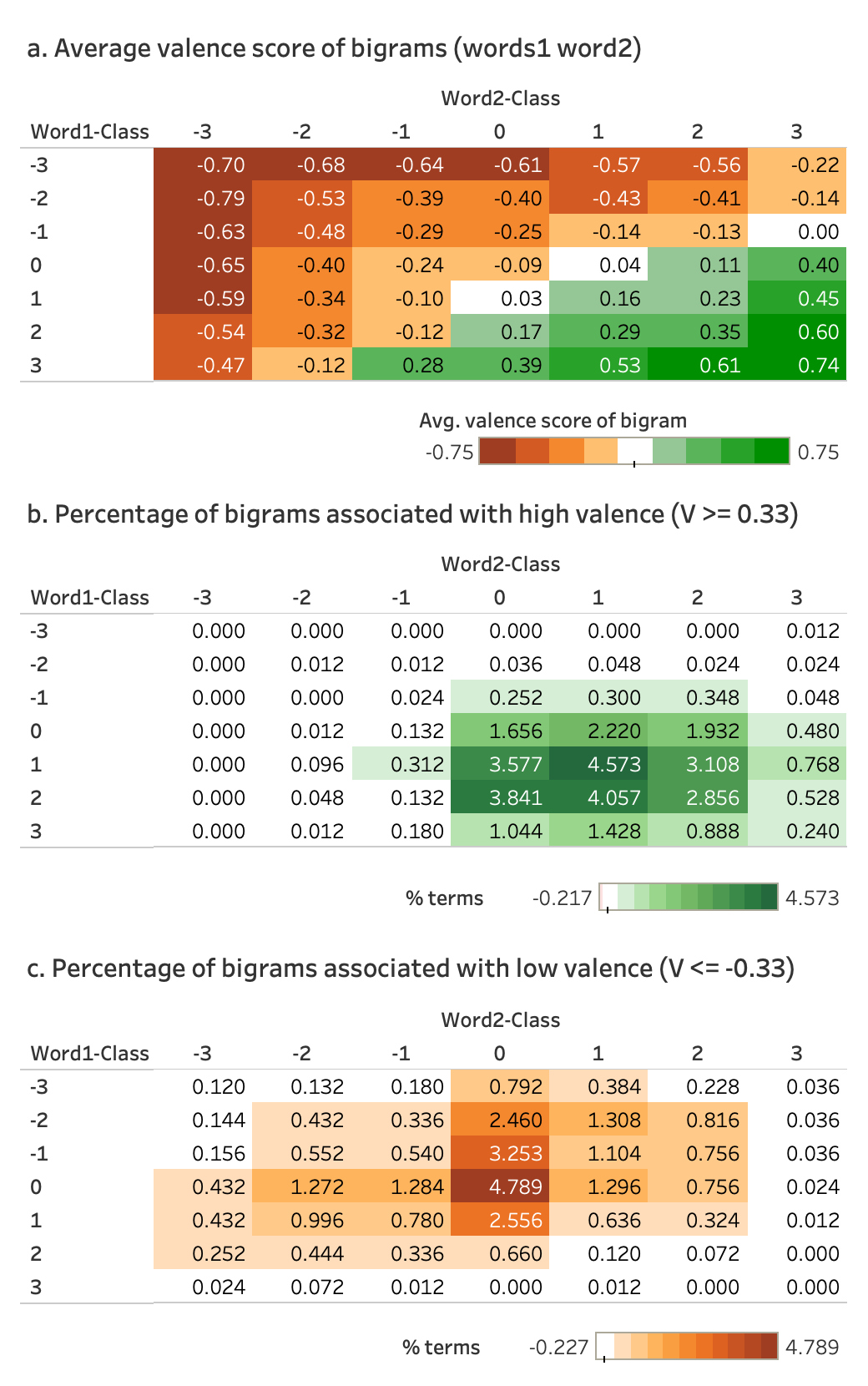}
        \vspace*{3mm}
        \caption{Measures of Valence Compositionality.}
        \vspace*{12mm}
	    \label{fig:v-compo}
\end{figure*}

\noindent \textbf{Results:} Observe in Figure \ref{fig:v-compo} (a) that as word1 and word2 (categorical) bin scores increase, the average valence score of the bigram bins also increases. 
This trend also exists for arousal and dominance (Figures \ref{fig:a-compo} (a), \ref{fig:d-compo} (a)), but it is markedly weaker.
The position of the higher V/A/D word (whether word1 or word2) does not seem to impact the scores much (corresponding scores on opposite sides of the diagonal are roughly similar).
Overall this suggests a marked degree of compositionality for all three dimensions, and more so for valence.

However, an examination of Figure \ref{fig:v-compo} (b) reveals that: the percentage of high-valence MWEs for which both word1 and word2 are neutral is not negligible (1.66\%); and markedly higher than the percentage for many other cells. 
Figure \ref{fig:v-compo} (c) shows that the percentage of low-valence MWEs for which both word1 and word2 are neutral is 4.79\% --- highest among all cells. Thus a large number of low-valence MWEs have neutral constituents; showing a high amount of noncompositionality. The trends for arousal and dominance are also similar (Figures \ref{fig:a-compo} and \ref{fig:d-compo}) showing that a large number of high and low A/D MWEs have neutral constituents.

Thus, overall, we conclude that a marked number of non-neutral MWEs are made up of neutral constituents; and that there is marked amount of emotional noncompositionality in MWEs (more so w.r.t.\@ arousal and dominance than valence).

\section{Applications and Future Work}

The large number of entries in the VAD Lexicon (for words and MWEs)  make it useful for a number of research inquiries and applications. We list a few below.\\[4pt]
\noindent  \textit{Especially relevant to MWEs}
% \begin{compactenum}
\begin{enumerate}
\item High- and low-valence MWEs reflect key psychological processes related to affect, motivation, memory, attention, and social communication. These MWEs like \textit{walk on air} (high valence) and \textit{rock bottom} (low valence) encode strong emotional meaning beyond the sum of their constituents. They can be used to study psychological processes such as reward processing, approach motivation, social bonding, resilience/coping, and self-evaluation.
    
\item High-arousal MWEs such as \textit{blow your mind}, \textit{burst into tears}, \textit{shake with rage}, and \textit{on edge}, often reflect, trigger, or describe intense physiological and psychological states. They can be used to study various psychological processes such as:\\[3pt]
Physiological Activation (Autonomic Arousal): states of heightened bodily activation (elevated heart rate, muscle tension, increased adrenaline, etc.). Examples: \textit{heart pounding} and \textit{on fire} (with excitement).\\[3pt]
Fight-or-Flight or Freeze Responses: Many MWEs metaphorically encode evolutionary survival responses, such as fight (aggression), flight (fear), or freeze (shock).
Examples:
\textit{Jump out of your skin} and \textit{Frozen with fear}.

\item High-dominance MWEs, such as \textit{take charge}, \textit{lay down the law}, \textit{crack the whip}, and \textit{have the upper hand}, reflect psychological processes tied to a strong sense of self-agency, where the subject is the initiator of actions and outcomes. Thus they can be useful in studying social power, assertiveness, threat readiness, and competence signaling.
In contrast, low-dominance MWEs, such as \textit{at the mercy of}, 
\textit{in over one’s head, thrown under the bus}, and \textit{toe the line},
reflect (and can be used to study) submission, helplessness, passivity, or external control. They often mark vulnerability or a lack of agency.

%  WASSA-2017 shared task on inferring the intensity of emotion felt by a person based on their tweet \cite{MohammadB17wassa}. 
% \item To study the  interplay between the categorical emotion model and the VAD model of affect. Much of the prior work has only explored one of the two models. The VAD lexicon can be used along with lists of words associated with emotions such as joy, sadness, fear, etc. to study the correlation of V, A, and D, with those emotions.\\[-20pt] 
% \item We will use the lexicon to identify syllables and phonemes that are associated with high V/A/D, that is, syllables that tend to occur more often than average in high V/A/D words. 

   \item 
   %\textit{Semantics and Human--Computer Interaction (HCI).}
MWEs such as \textit{over the moon}, \textit{make someone's day, out of sorts}, and \textit{keep your cool}  have non-literal meanings that cannot be determined simply from the meanings of their constituents.
They can be especially revealing in how people store and retrieve language chunks of meaning.
Thus, both linguists and developers of artificial chat agents benefit from a large repository of MWE--VAD associations. 

\item %\textit{Pragmatics, Emotions, and Framing.}
MWEs are often laden with rich connotative meaning, conveying subtle nuances of emotional intensity, politeness, formality, or social acceptability \cite{sag2002multiword,zgusta1967multiword,citron2019idiomatic,allawama2025idioms}. MWEs such as \textit{losing control, on edge,} \textit{on cloud nine}, \textit{at peace with myself}, or \textit{on top of the world} encode folk psychological concepts of emotional states.
These expressions reveal how people naturally talk about and categorize emotion, which helps researchers build models of affect.
MWEs often have an emotional punch, thereby influencing perception, recall, and judgment \cite{citron2019idiomatic}.
Thus, MWEs like \textit{war on drugs} and \textit{family values} are used to frame complex issues in persuasive ways. Therefore, MWEs, especially those associated with high and low V/A/D, are highly relevant to studying discourse analysis and political rhetoric.

\item % \textit{Cognitive Science, Cognitive Linguistics, Cognitive Psychology.}
MWEs often draw on physical and embodied metaphors \cite{kacinik2014sticking}. Examples of emotional and embodied MWEs include: 
\textit{feeling down}	(valence is verticality: low valence is low and high valence is high),
glow with happiness	(valence is degree of light and high valence is bright whereas low valence is dark)
boil with rage	(anger is heat and strong arousal raises internal temperature),
running on empty (body is a machine),
step up	(power is elevation), etc.
Thus MWEs are a window into embodied and metaphorical thinking. 
MWEs associated with high and low V/A/D  can be used to study how emotional language is grounded in sensorimotor experience and is organized metaphorically.

\item % \textit{Affective Science.}
MWEs frequently reflect emotion regulation strategies, both maladaptive and adaptive \cite{nichter2010idioms,lee2017figurative,cole2010role}. Examples of  MWEs pertaining to emotion regulation, include: \textit{bottling it up, trying to push it down, taking a deep breath,} and  \textit{letting it go}.
They give insight into implicit self-regulatory processes people engage in during emotional episodes.
MWEs provide linguistic evidence for how affect interacts with attention, memory, appraisal, and prediction (core components of emotion theories).
MWE-VAD can be used to study emotion regulation strategies and inform theories of emotion by showing how people encode appraisals and attention patterns in everyday language.
\end{enumerate}
% \end{compactenum}

% \newpage
\noindent \textit{Relevant Generally (to Words and MWEs)}

\begin{compactenum}

\item  Understanding valence, arousal, and dominance, and the underlying mechanisms; how VAD relate to our mind and body; how VAD change with age, socio-economic status, weather, green spaces, etc.
\item Determining how VAD manifest in language; how language shapes our VAD; how culture shapes the language of VAD; etc.
\item Tracking the degree of VAD towards targets of interest such as climate change, %BB commercial products, 
government policies, biological vectors, etc. %; tracking common targets of anxiety;  
% \item Studying stereotypes and social cognition; using the dominance aka competence lexicon to study how competence assessment capabilities develop in children and to track perceptions of competence towards various targets of interest.
\item Developing automatic systems for detecting VAD; To provide features for automatic sentiment or emotion detection systems. They can also be used to obtain sentiment-specific word embeddings and sentiment-specific sentence representations.
% \item MWE-VAD can be used to identify syllables and sub-word units that consistently tend to occur in words with high VAD scores. This has implications in understanding how some syllables and sounds have a tendency to occur in words referring to semantically related concepts. Identifying V, A, and D scores associated with syllables is also useful in generating names for literary characters and commercial products that have the desired affectual response.  
\item MWE-VAD can be used to study emotions in story telling; its relationship with central elements of narratology such as conflict and resilience.
To identify high V, A, and D words and MWEs in books and literature. To facilitate work of researchers in digital humanities. To facilitate work on literary analysis.
\item MWE-VAD is a source of gold (reference) scores, the entries in the VAD lexicon can be used in the evaluation of automatic methods of determining word--VAD associations.
% \item The VAD lexicon is also of potential use to psychologists and evolutionary linguists interested in determining how evolution shaped the representation of the world around us, and why certain personality traits are associated with higher or lower shared understanding of valence, arousal, and dominance of words.
    
\end{compactenum}
\noindent Thus language resources at the intersection of MWEs and VAD are highly relevant to our understanding of a wide variety of phenomena.
\bl{Note that automatic prediction of valence/sentiment/emotions from individual text instances is only a small part of the use cases. The lexicon can be used to obtain new insights on a wide variety of research questions (including those that are most directly answered by the lexicon rather than by using some ML system or LLM). Finally, even though large language models can at times be used in place of lexicons, any inferences drawn from an automatic approach requires manual validation. Portions of the manually created VAD lexicon presented here can be used to improve the generations of the LLM and held out portions can be used to validate the LLM generations.}

% \noindent Apart from exploring the applications above, we are also interested in creating VAD lexicons for other languages, especially Chinese, Hindi, Arabic, Spanish, and German. We can then explore characteristics of valence, arousal, and dominance that are common across cultures. 

\section{Conclusions}
We present here the MWE-VAD Lexicon, which has human ratings of valence, arousal, and dominance for more than
% than 44,000 English words and 10,000 multi-word expressions. 
10,000 English MWEs and 25,000 unigrams.
Notably the 25k unigrams are words not included in the NRC VAD Lexicon v1, and so greatly increasing coverage for unigrams.
We show that the ratings are highly reliable
(split-half reliability of over 0.95 for all three dimensions). 
\bl{We add these entries to those in NRC VAD Lexicon v1 to create v2.}
We use the lexicon to study the 
% distribution of various VAD classes in different 
\bl{the extent to which different MWE types express strong emotionality}. We also quantify the degree of emotional compositionality of MWEs with various metrics.
Finally, we make a case for why language resources of MWEs associated with valence, arousal, and dominance are useful for a wide array of research inquiries and applications
% rate at which children acquire VAD words with age. 
% Finally, we show that 
% The lexicon enables a wide variety of  %bias and stereotype 
% research 
in Psychology, NLP, Public Health, Digital Humanities, and Social Sciences. 
\bl{The NRC VAD Lexicon v2 is freely available for research 
through the project webpage.}\footnote{\url{http://saifmohammad.com/WebPages/nrc-vad.html}}

\section{Limitations}
\label{sec:limitations}

The lexicon created is one of the largest that exist with wide coverage and a large number of annotators (thousands of people as opposed to just a handful). 
However, no lexicon can cover the full range of linguistic and cultural diversity in emotion expression. 
The lexicons are restricted to words that are most commonly used in Standard American English and they capture emotion associations as judged by American native speakers of English. 
% We hope that future work will build similar lexicons with participants from various other regions and for various other languages.
Annotators on Mechanical Turk are not representative of the wider US population. However, obtaining annotations from a large number of annotators (as we do) makes the lexicon more resilient to individual biases and captures more diversity in beliefs.
We see this work as a first step that paves the way for more work using responses from various other groups of people and in various other languages. 
We built our lexicon using many of the principles and ideas listed in \citet{Mohammad23ethicslex}, which provides a detailed discussion of the limitations and best-practices in the creation and use of emotion lexicons.

% \noindent See discussions of limitations in how the lexicons can be used and interpreted in the Ethics Statement below (\S \ref{sec:ethics}).

\section{Ethics and Data Statement}
\label{sec:ethics}

The crowd-sourced task presented in this paper was approved by our Institutional Research Ethics Board. 
 The individual words and MWEs selected did not pose any risks beyond the risks of occasionally reading text on the internet. 
The annotators were free to do as many word and MWE annotations as they wished. The instructions included a brief description of the purpose of the task (Figures \ref{fig:val-q} through \ref{fig:dom-ex}}).

VAD assessments are complex, nuanced, and often instantaneous mental judgments. Additionally, each individual may use language to convey these assessments slightly differently.
\bl{See \citet{Mohammad23ethicslex} for a discussion of good practices and ethical considerations when using emotion lexicons. See \citet{Mohammad22AER} for a broader discussion of ethical considerations relevant to automatic emotion recognition.
We discuss below notable points of discussion as well as some new and updated points especially relevant for VAD and MWE norms.}
% See \citet{Mohammad23ethicslex} for a discussion of good practices and ethical considerations when using emotion lexicons. See \citet{Mohammad22AER} for a broader discussion of ethical considerations relevant to automatic emotion recognition.
% See \citet{Mohammad23ethicslex} for a detailed discussion of ethical considerations when computationally analyzing emotions and VAD using emotion lexicons. 
% We discuss below some notable ethical considerations when using our lexicon. 
%  (See \citet{Mohammad22AER} for a broader discussion of ethical considerations relevant to automatic emotion recognition.)
\begin{compactenum}
    \item \textit{Coverage:} We sampled a large number of English words from other lexical sources (which themselves sample from many sources). Yet, the words included do not cover all domains, genres, and people of different locations, socio-economic strata, etc.\@ equally. It likely includes more of the vocabulary and MWEs used by people in the United States and with socio-economic and educational backgrounds that allow for technology access.
 
    \item \textit{Word Senses and Sense Priors:} Words when used in different senses and contexts may be associated with different degrees of VAD associations. The entries in in the VAD Lexicon are indicative of the associations with the predominant senses of the words. This is usually not problematic because most words have a highly dominant main sense (which occurs much more frequently than the other senses). 
In specialized domains, some terms might have a different dominant sense than in general usage. Entries in the lexicon for such terms should be appropriately updated or removed. 
Further, any conclusions using the lexicon should be made based on relative change of associations using a large number of textual tokens. For example, if there is a marked increase in low-valence words from one period to the next, where each period has thousands of word tokens, then the impact of word sense ambiguity is minimal, and it is likely that some broader phenomenon is causing the marked increase in low-valence words. (See last two bullets.)

    \item \textit{Not Immutable:} The VAD scores do not indicate an inherent unchangeable attribute. The associations can change with time (e.g., the decrease in negativeness associated with \textit{inter-race relationships} over the last 100 years), but the lexicon entries are fixed. They pertain to the time they are created. However, they can be updated with time.
  
    \item \textit{Socio-Cultural Biases:} \bl{Many multiword expressions have origins and connotations in historic racism and bigotry, e.g., {\it sold down the river, grandfathered in,} and {\it black sheep}. Many have argued that, in everyday speech, choosing alternative expressions fosters more inclusiveness. On the other other hand, use of such expressions in research can shed light on the historical and social context of racism and stereotypes permeate language.  
    }
    
    The annotations for VAD capture various human biases. These biases may be systematically different for different socio-cultural groups. Our data was annotated by mostly US
    % , Canadian, UK, and Indian 
    English speakers, but even within a country there are many diverse socio-cultural groups.
    Notably, crowd annotators on Amazon Mechanical Turk do not reflect populations at large. In the US for example, they tend to skew towards male, white, and younger people. However, compared to studies that involve just a handful of annotators, crowd annotations benefit from drawing on hundreds and thousands of annotators (such as this work). 
 
    % \item \textit{Inappropriate Biases:} 
    % Our biases impact how we view the world, and some of the biases of an individual may be inappropriate. For example, one may have race or gender-related biases that may percolate subtly into one's notions of VAD associated with words. 
    Our dataset curation was careful to avoid words and MWEs from problematic sources. We also asked people annotate terms based on what most English speakers think (as opposed to what they themselves think). This helps to some extent, but the lexicon may still capture some historical VAD associations with certain identity groups. This can  be useful for some socio-cultural studies; but we also caution that VAD associations with identity groups be carefully contextualized. % to avoid false conclusions.    
  
    \item \textit{Perceptions (not “right” or “correct” labels):} Our goal here was to identify common perceptions of WTS association. These are not meant to be ``correct'' or ``right'' answers, but rather what the majority of the annotators believe based on their intuitions of the English language.
   
    \item \textit{Avoid Essentialism:} When using the lexicon alone, it is more appropriate to make claims about VAD word usage rather than the VAD of the speakers. For example, {\it `the use of high-valence words in the context of the target group grew by 20\%'} rather than {\it `valence in the target group grew by 20\%'}. In certain contexts, and with additional information, the inferences from word usage can be used to make broader VAD claims. 
\item \textit{Avoid Over Claiming:} Inferences drawn from larger amounts of text are often more reliable than those drawn from small amounts of text.
 For example, {\it `the use of high-valence words grew by 20\%'} is informative when determined from hundreds, thousands, tens of thousands, or more instances. Do not draw inferences about a single sentence or utterance from the VAD associations of its constituent words.
\item \textit{Embrace Comparative Analyses:} Comparative analyses can be much more useful than stand-alone analyses. Often, VAD word counts and percentages on their own are not very useful. 
For example, {\it `the use of high-valence words grew by 20\% when compared to [data from last year, data from a different person, etc.]'} is more useful than saying {\it `on average, 5 high-valence words were used in every 100 words'}.
\end{compactenum}
\noindent We recommend careful reflection of ethical considerations relevant for the specific context of deployment when using the VAD lexicon.

% \section*{Acknowledgments}

% Many thanks to Tara Small for  helpful discussions.

%\bibliography{acl2017}
% \section{Bibliographical References}
\bibliography{anthology,maxdiff}

\appendix

\section{APPENDIX}
\label{appendix-a}

\subsection{AMT Questionnaires for Valence, Arousal, and Dominance}
\begin{figure*}[t]
	     \centering
	     \includegraphics[width=\textwidth]{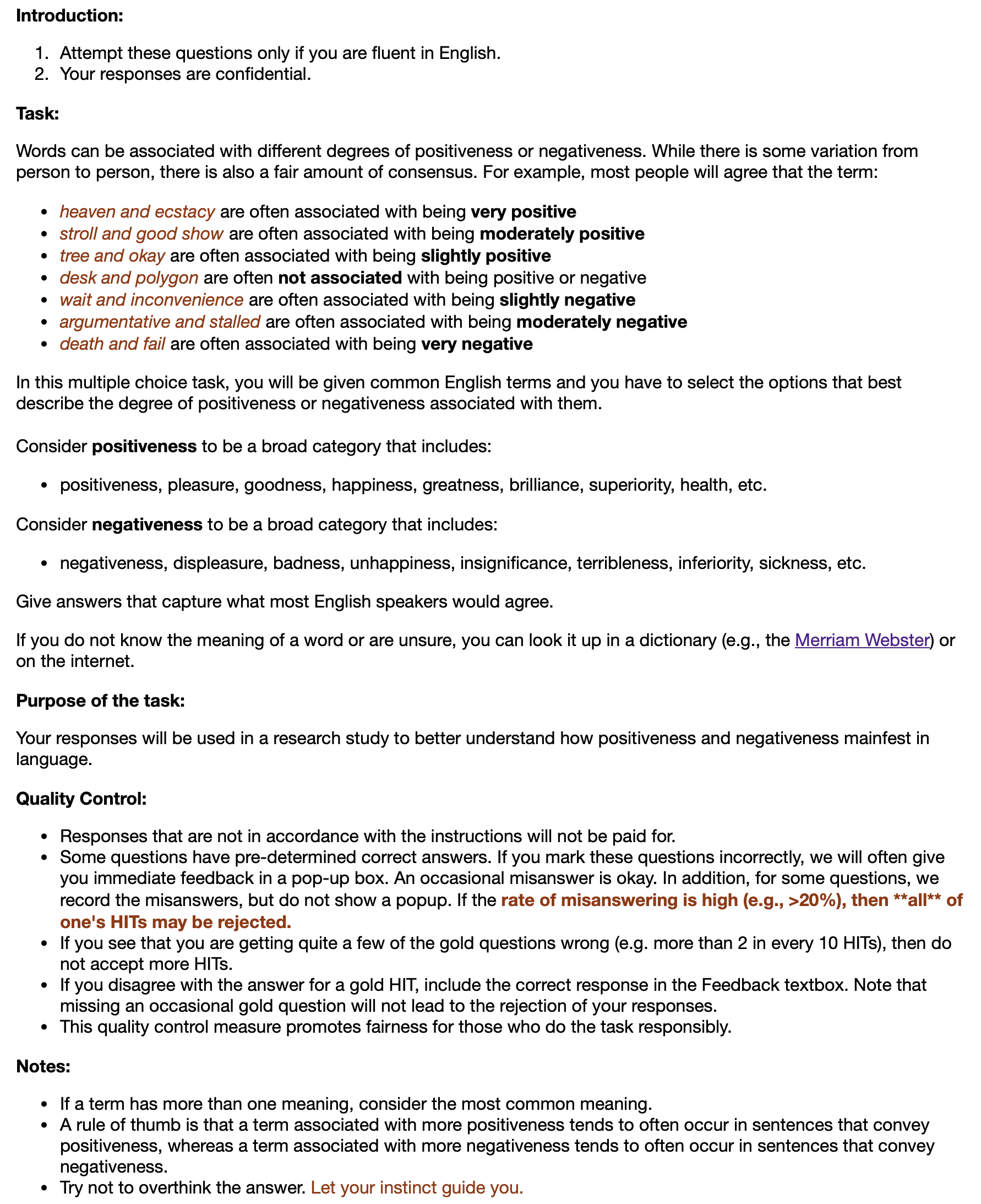}
	     \caption{Valence Questionnaire: Detailed instructions.}
	     \label{fig:val-q}
	 \end{figure*}

\begin{figure*}[t]
	     \centering
	     \includegraphics[width=\textwidth]{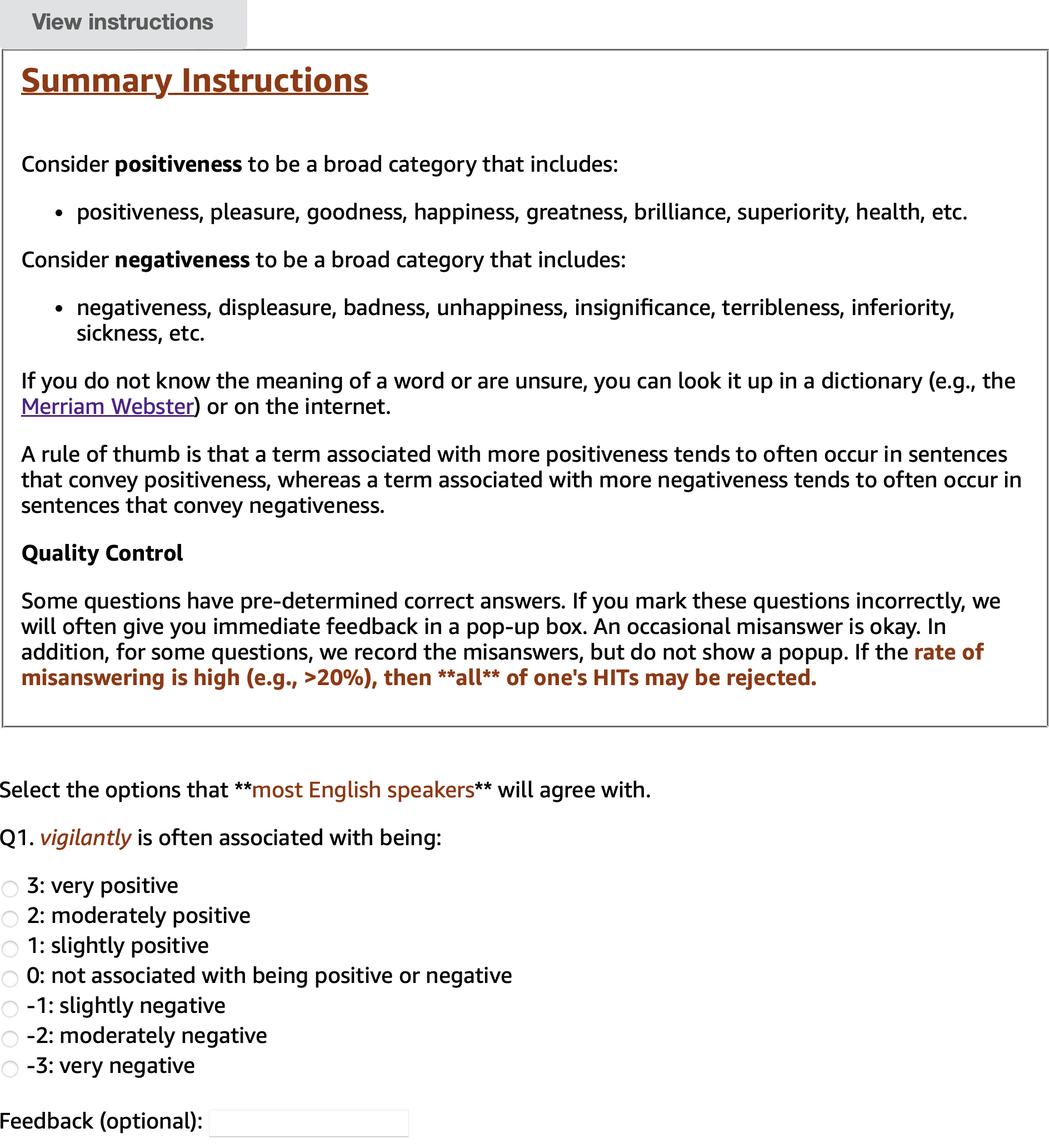}
	     \caption{Valence Questionnaire: Sample question.}
	     \label{fig:val-sumq}
	 \end{figure*}

\begin{figure*}[t]
	     \centering
	     \includegraphics[width=0.65\textwidth]{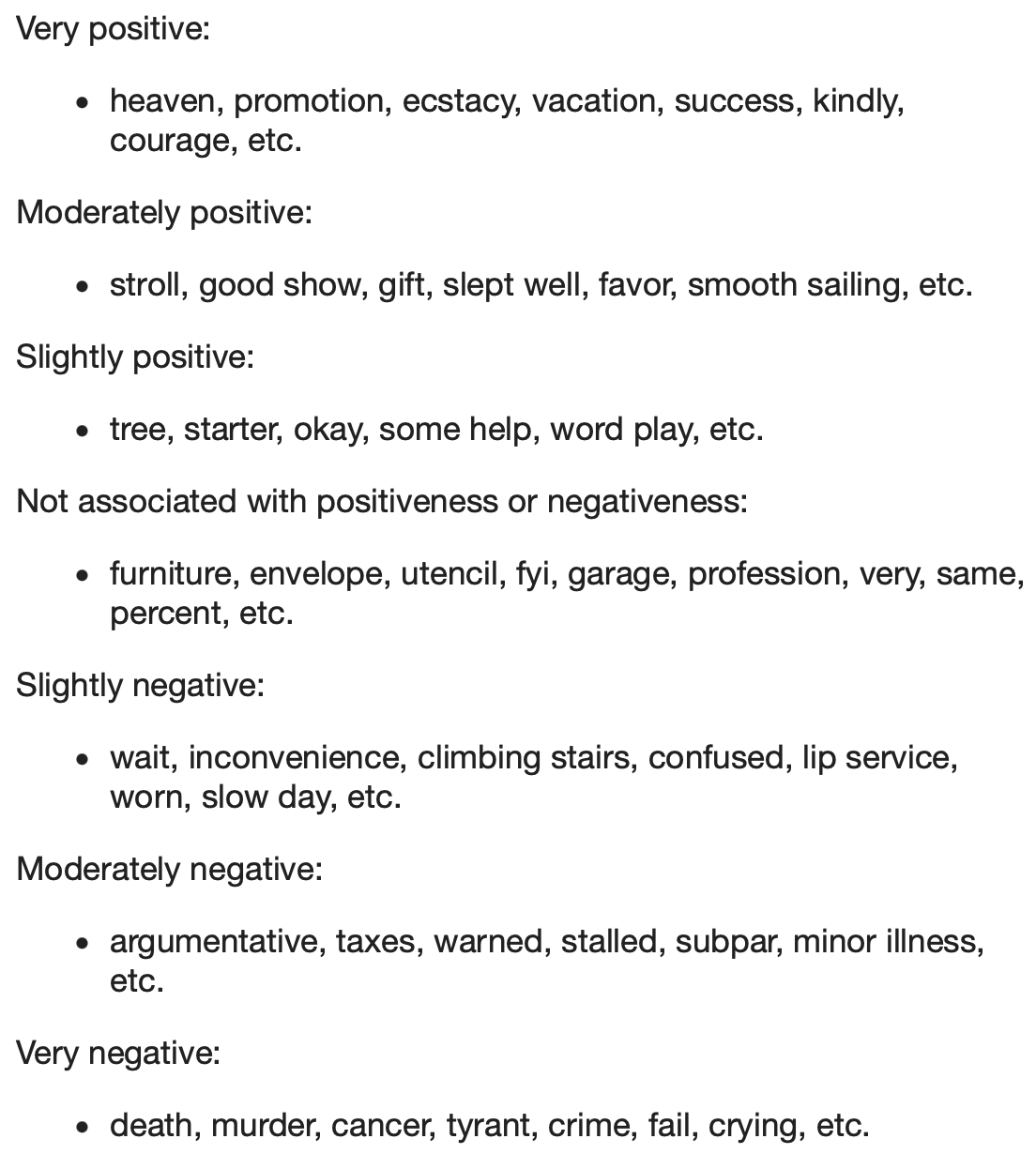}
	     \caption{Valence Questionnaire: Examples.}
	     \label{fig:val-ex}
	 \end{figure*}

\begin{figure*}[t]
	     \centering
	     \includegraphics[width=\textwidth]{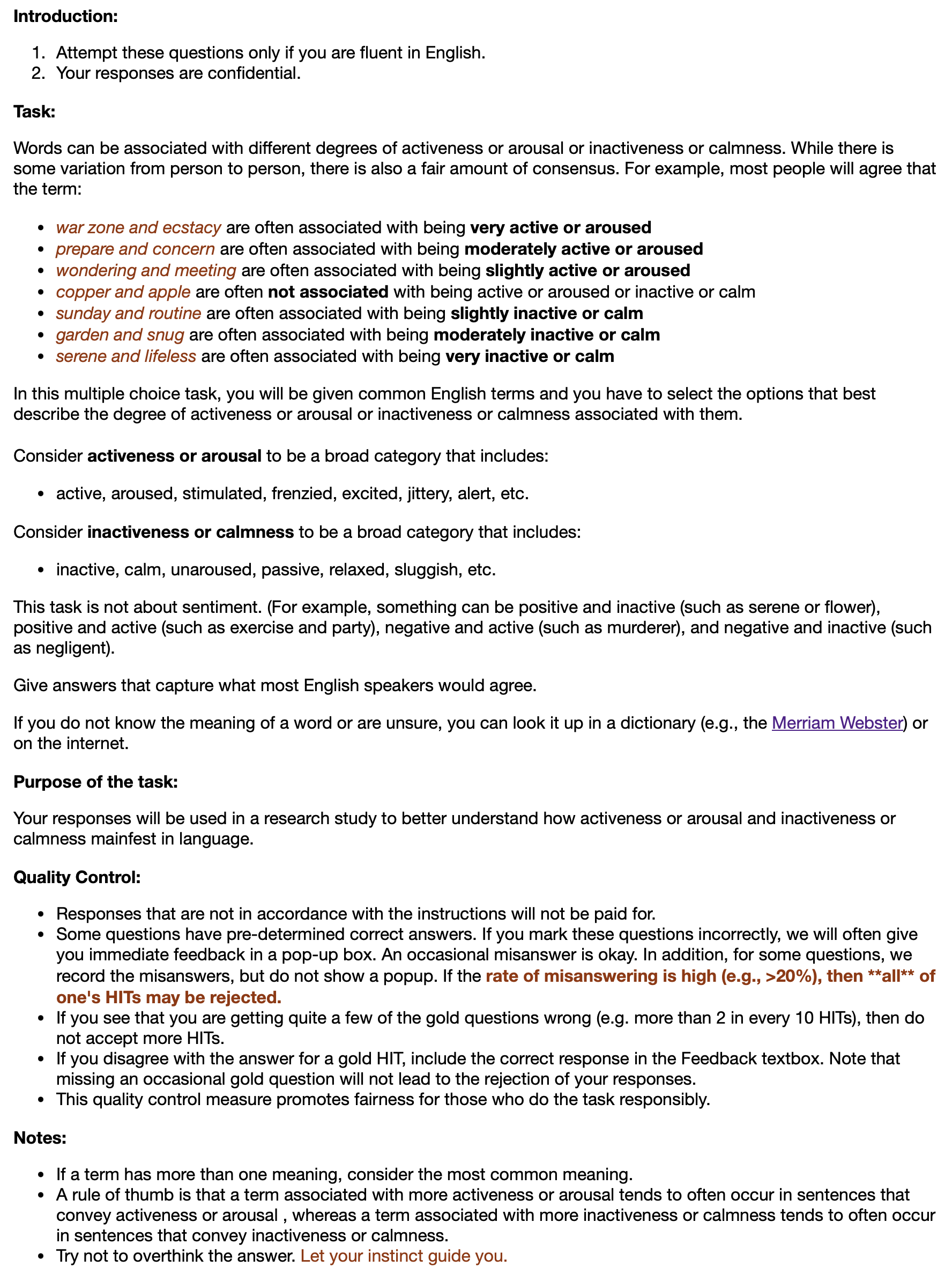}
	     \caption{Arousal Questionnaire: Detailed instructions.}
	     \label{fig:aro-q}
	 \end{figure*}

\begin{figure*}[t]
	     \centering
	     \includegraphics[width=\textwidth]{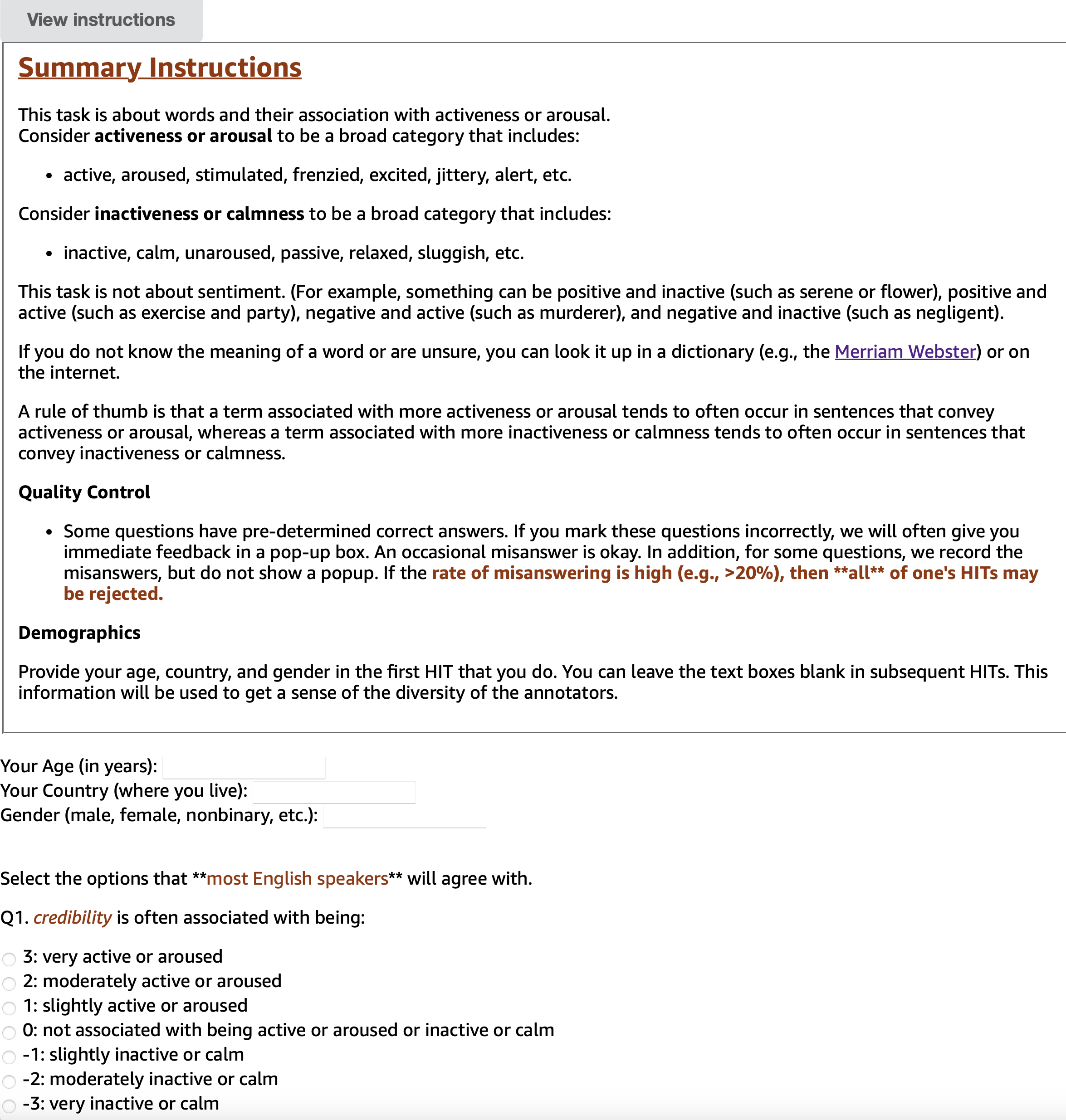}
	     \caption{Arousal Questionnaire: Sample question.}
	     \label{fig:aro-sumq}
	 \end{figure*}

\begin{figure*}[t]
	     \centering
	     \includegraphics[width=0.75\textwidth]{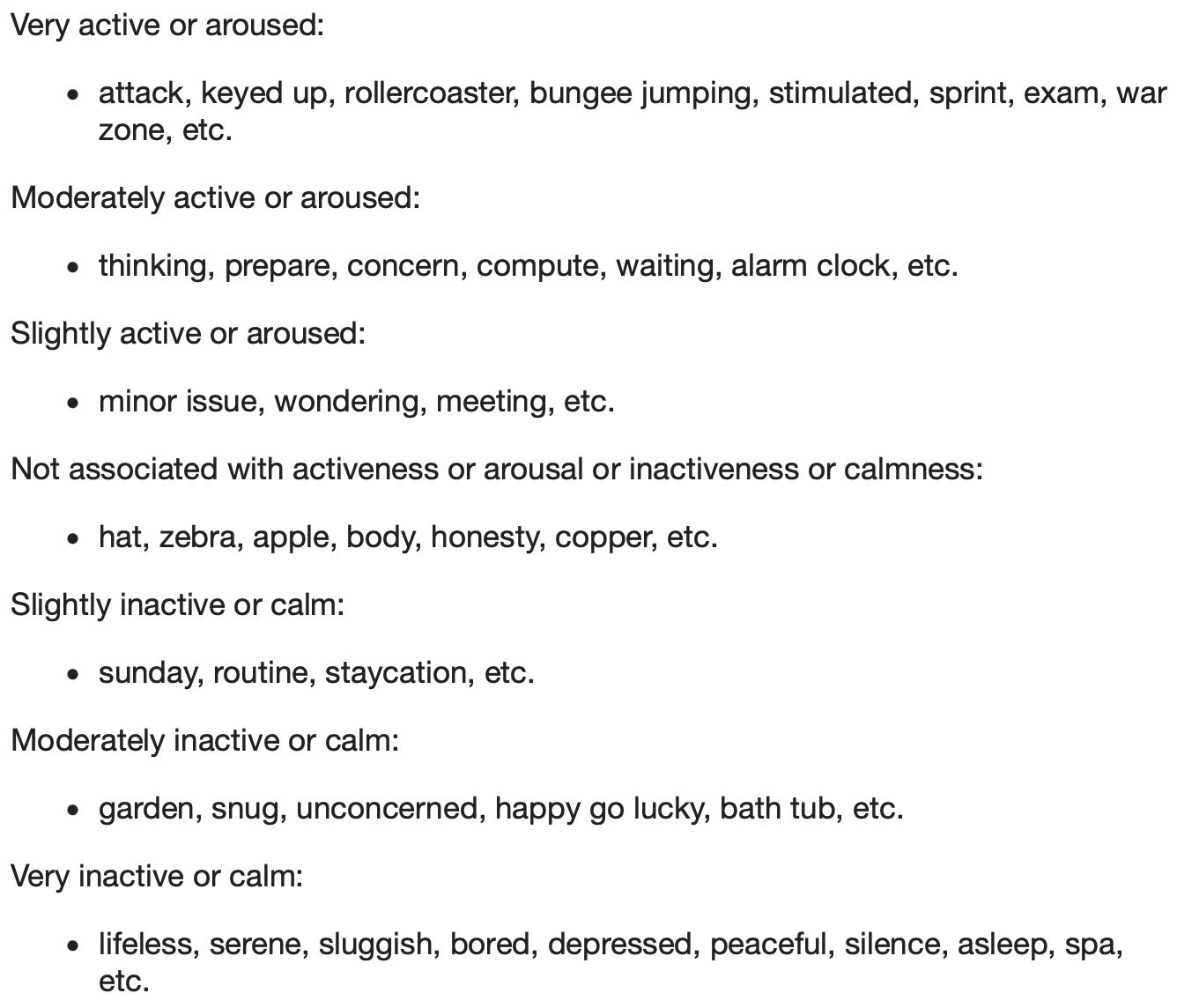}
	     \caption{Arousal Questionnaire: Examples.}
	     \label{fig:aro-ex}
	 \end{figure*}

\begin{figure*}[t]
	     \centering
	     \includegraphics[width=\textwidth]{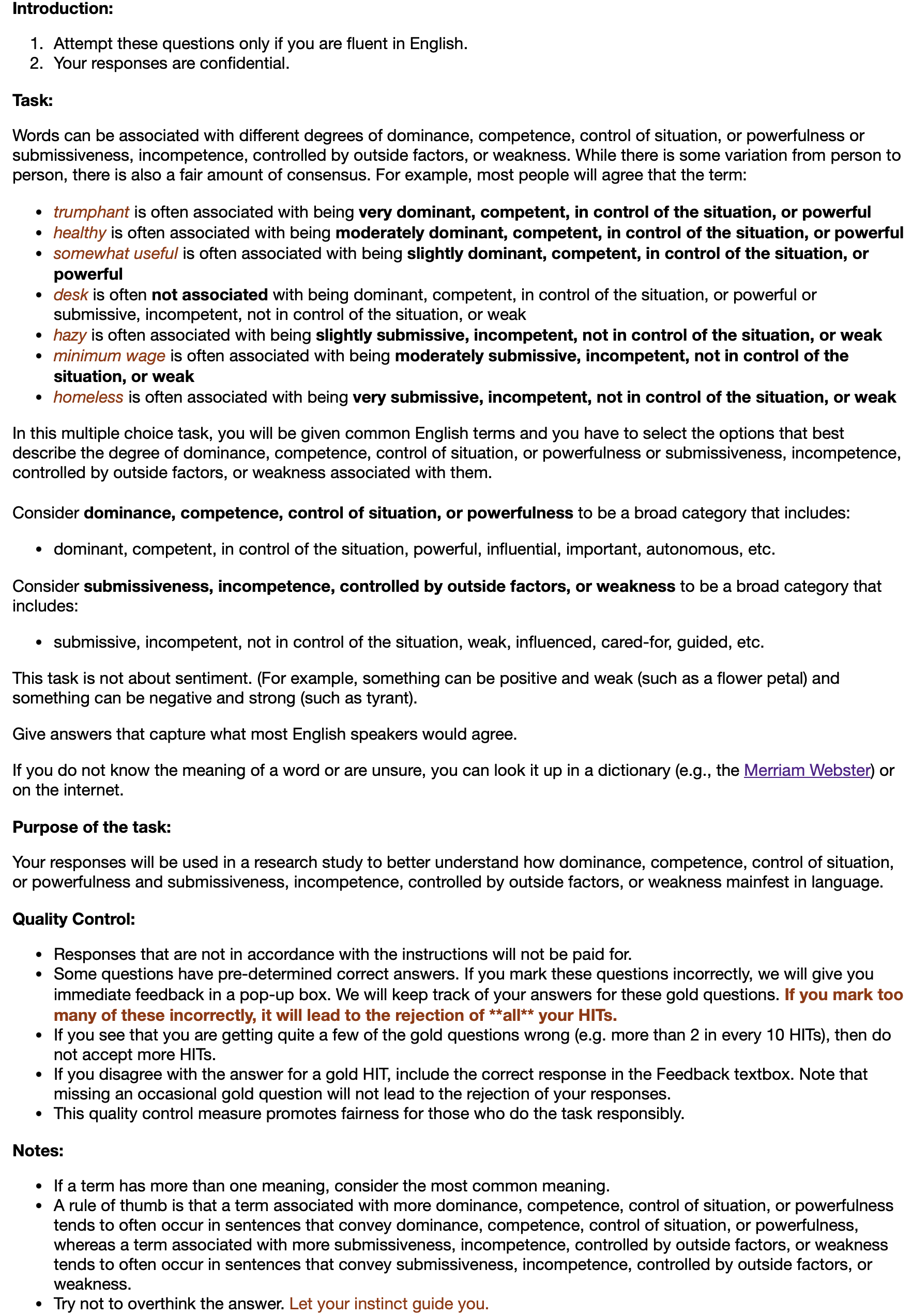}
	     \caption{Dominance Questionnaire: Detailed instructions.}
	     \label{fig:dom-q}
	 \end{figure*}

\begin{figure*}[t]
	     \centering
	     \includegraphics[width=\textwidth]{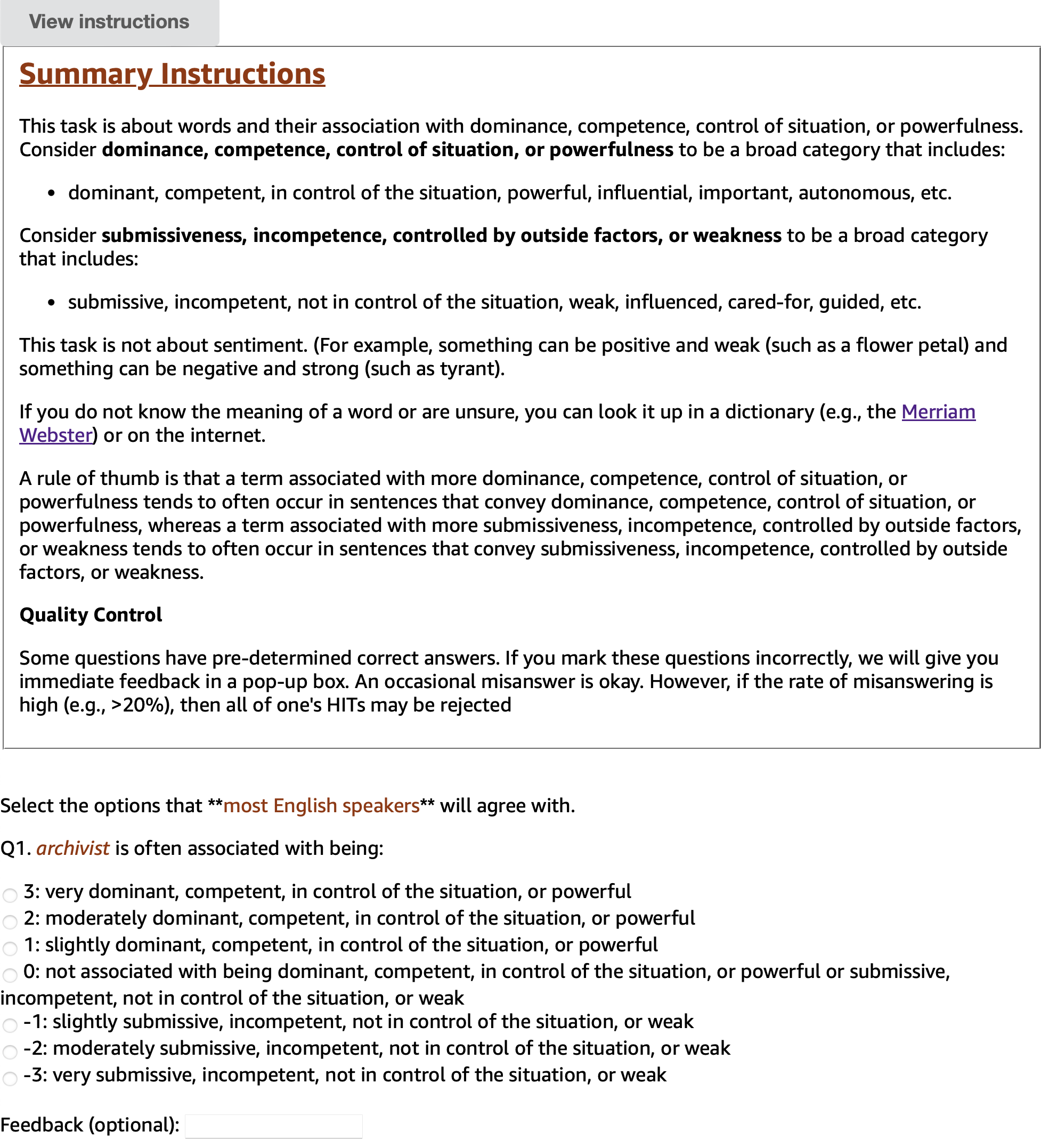}
	     \caption{Dominance Questionnaire: Sample question.}
	     \label{fig:dom-sumq}
	 \end{figure*}

\begin{figure*}[t]
	     \centering
	     \includegraphics[width=0.75\textwidth]{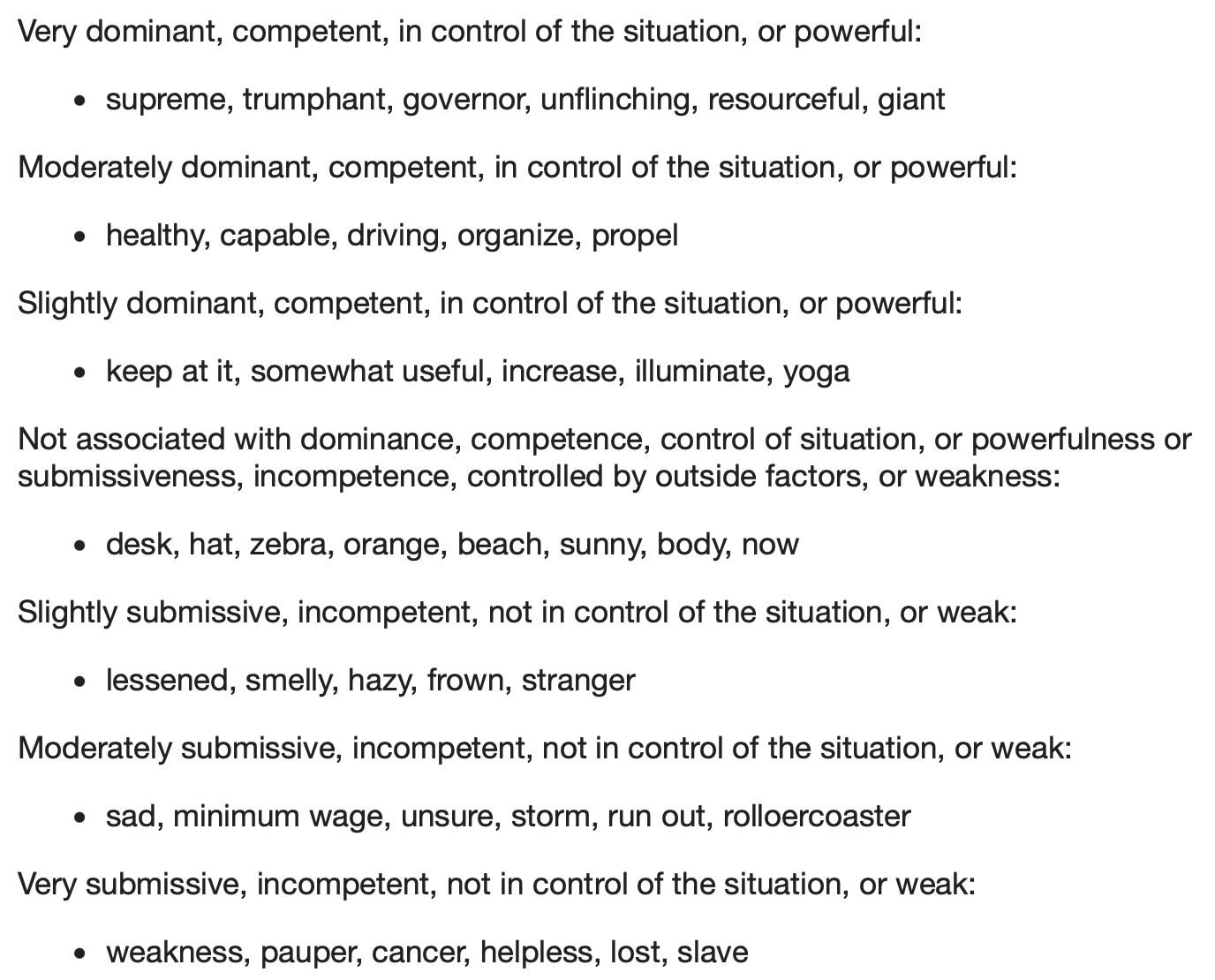}
	     \caption{Dominance Questionnaire: Examples.}
	     \label{fig:dom-ex}
	 \end{figure*}

Screenshots of the detailed instructions, sample instance (question), and examples presented to the annotators are shown in Figures \ref{fig:val-q} through \ref{fig:dom-ex}. Participants were informed that they may work on as many instances as they wish. 
The annotation task was approved by our institution's IRB. 
The purpose of the task and how their annotations will be used was made clear, and consent was obtained.

\subsection{Distribution of MWE-VAD}

MWE-VAD is made freely available on the project website as a compressed file. Terms of use will require that users not re-distribute the file and not post any form of the lexicon on the web. This is to prevent the resource being included in the data scrape fed to a large language model. 
See full list of terms of use at the project home page.

\subsection{Computational Resources and Carbon Footprint}
A nice advantage of using simple lexicon-based approaches is the low carbon footprint and computational resources required. All of the experiments described in the paper were conducted on a regular personal laptop.

% ----------------

% ----------------

\section{Supplementary Figures}
\begin{figure}[t!]
	\centering
	    \includegraphics[width=0.47\textwidth]{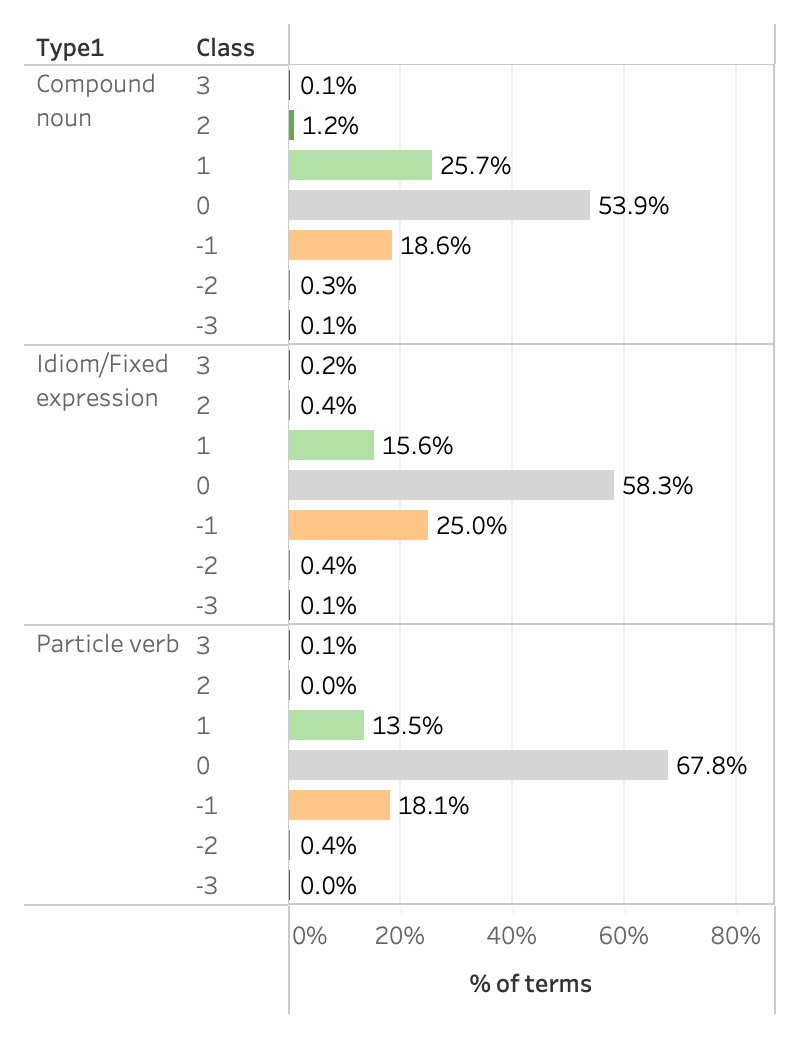}
        \caption{Percentage of MWEs pertaining to each of the \textbf{arousal} classes within each MWE type.}
        \vspace*{-3mm}
	    \label{fig:aro-distrib}
\end{figure}

\begin{figure}[t!]
	\centering
	    \includegraphics[width=0.47\textwidth]{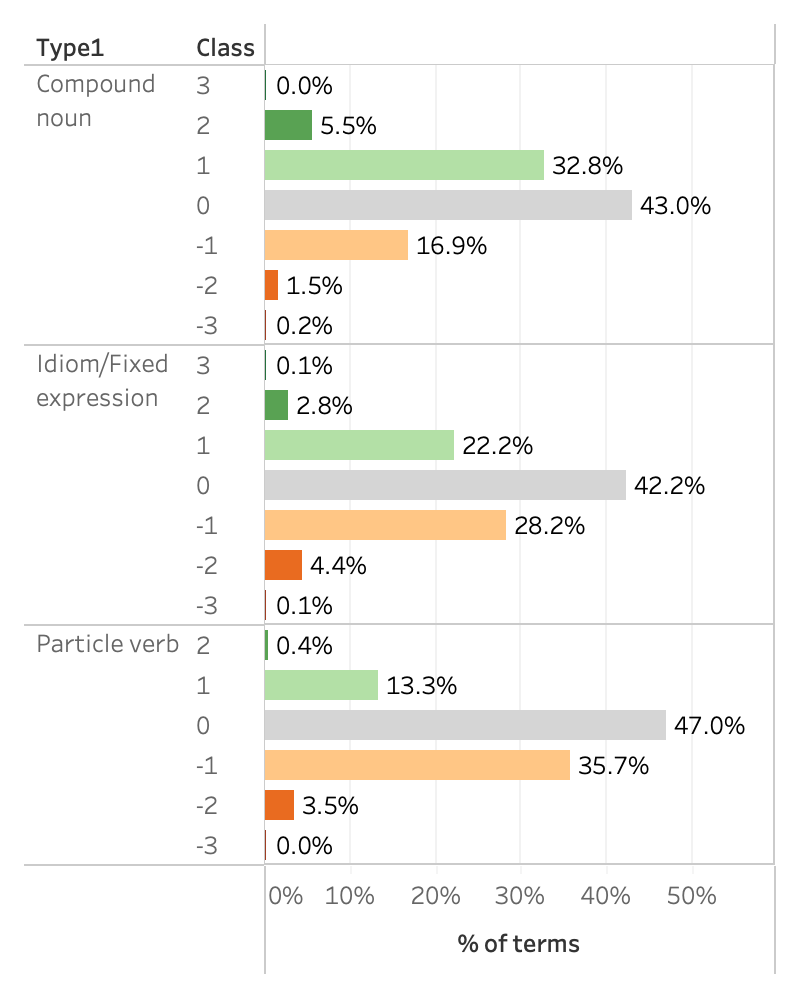}
        \caption{Percentage of MWEs pertaining to each of the \textbf{dominance} classes within each MWE type.}
        \vspace*{-3mm}
	    \label{fig:dom-distrib}
\end{figure}

\begin{figure*}[t!]
	\centering
	    \includegraphics[width=0.72\textwidth]{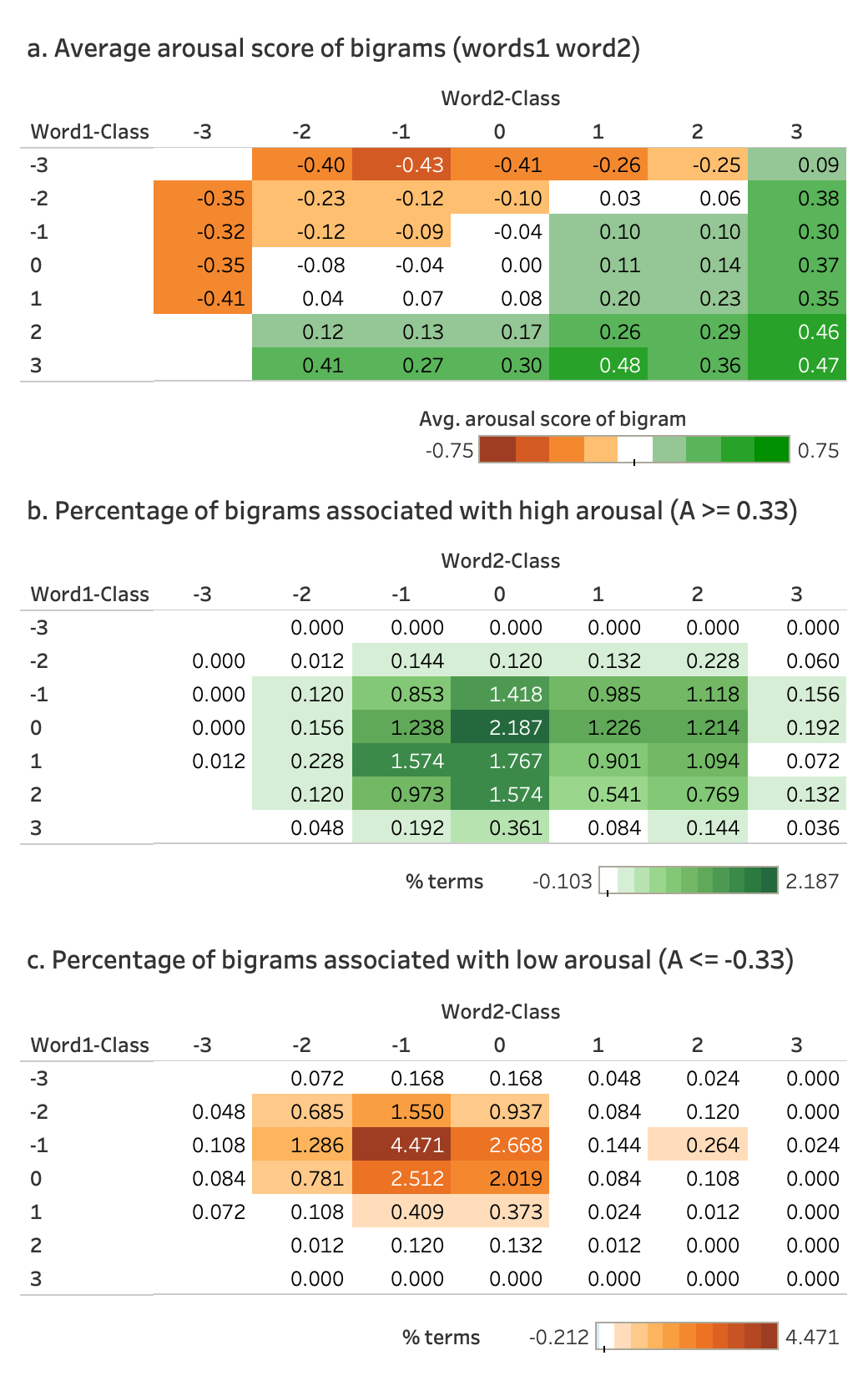}
        \caption{Measures of Arousal Compositionality.}
        \vspace*{-3mm}
	    \label{fig:a-compo}
\end{figure*}

\begin{figure*}[t!]
	\centering
	    \includegraphics[width=0.72\textwidth]{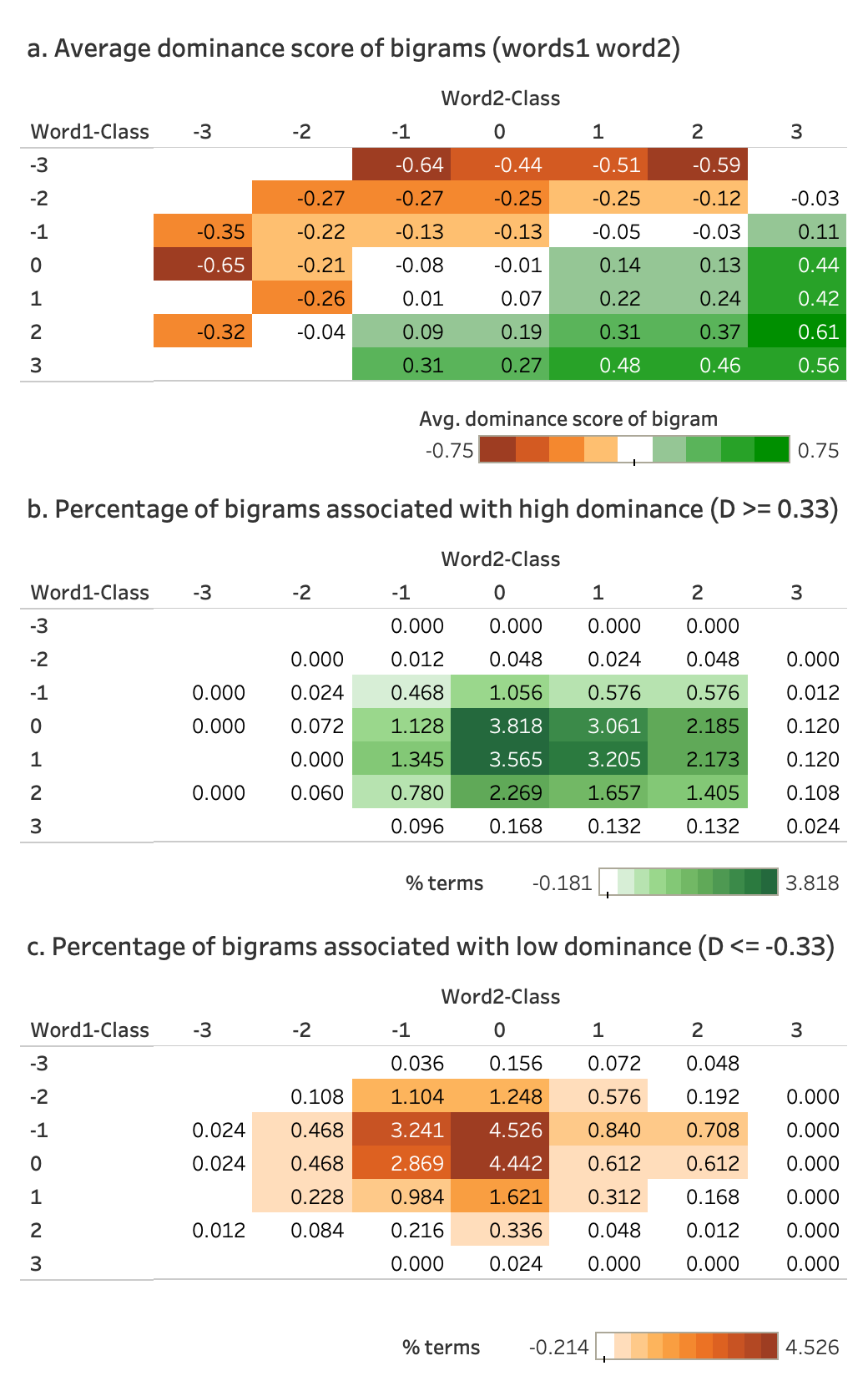}
        \caption{Measures of Dominance Compositionality.}
        \vspace*{-3mm}
	    \label{fig:d-compo}
\end{figure*}

Distributions of the arousal and dominance classes in MWE types are shown in Figures \ref{fig:aro-distrib} and  \ref{fig:dom-distrib}.
Metrics for arousal and dominance compositionality are shown in Figures 
\ref{fig:a-compo} and \ref{fig:d-compo}.

\end{document}